\documentclass{anstrans}

\usepackage{graphicx}
\usepackage{microtype}
\usepackage{paralist}
\usepackage{caption}
\usepackage{makecell}
\usepackage{subcaption}

\title{Aggregation of Published Non-Uniform Axial Power Data for Phase II of the OECD/NEA AI/ML\\Critical Heat Flux Benchmark}

\author{Reece~Bourisaw$^{1}$, Reid McCants$^{1}$, Jean-Marie Le Corre$^{2}$, Anna~Iskhakova$^{1}$ and Arsen~S.~Iskhakov$^{1}$}

\institute{$^{1}$Department of Mechanical and Nuclear Engineering, Kansas State University, Manhattan, KS, \and $^{2}$Westinghouse Electric Sweden AB, 72163 V\"aster\aa s, Sweden}

\email{rbourisaw@ksu.edu}

\doi{doi.org/10.13182/T133-49117}

\begin{document}
\section{Introduction}
Critical heat flux (CHF) is a key thermal‐hydraulic phenomenon in which the two‐phase coolant's heat‐transfer capability drops abruptly, establishing the thermal‐design limit, defined as the departure from nucleate boiling (DNB) in pressurized water reactors (PWRs) and as the critical‐power condition in boiling water reactors (BWRs)~\cite{todreas2021}. Therefore, accurate prediction of CHF is essential for setting safe operating margins and maximizing reactor power output without risking fuel damage. However, CHF modeling remains challenging due to the complex coupling of local two‐phase hydrodynamics, surface characteristics, transient thermal boundary‐layer development, and geometric influences, factors that vary widely with pressure, flow rate, and heating profile~\cite{lee1988,guo2025}.

Traditional CHF correlations often rely on simplified mechanistic assumptions, empirical parametric relationships or look‐up tables (LUT), which can limit their accuracy and adaptability under novel or strongly non-uniform heating conditions~\cite{groeneveld1986,groeneveld2007}. Machine learning (ML) methods, by contrast, can learn high‐dimensional patterns directly from large datasets and have demonstrated improved accuracy and robustness in CHF prediction~\cite{grosfilley2024}. However, many ML‐based CHF studies have been carried out independently as proof‐of‐concept demonstrations on limited or proprietary datasets, underscoring the need for coordinated benchmarking efforts. To address this, the Organization for Economic Co-operation and Development / Nuclear Energy Agency (OECD/NEA) Task Force on Artificial Intelligence and Machine Learning for Scientific Computing in Nuclear Engineering organized a CHF benchmark to provide common datasets, blind‐test partitions, and reproducible evaluation protocols for evaluating ML‐based CHF predictors~\cite{lecorre2024}.

Phase I of the CHF benchmark, as specified in~\cite{lecorre2024}, comprised four tasks: (1)~dimensionality analysis for feature selection and extraction; (2)~development and optimization of ML‐based CHF regression models based on the Nuclear Regulatory Commission (NRC) database; (3)~rigorous model evaluation and uncertainty quantification (UQ); and (4)~independent blind testing on withheld data. While Phase~I demonstrated the feasibility of AI/ML approaches for CHF prediction under uniform heating, it remained limited by the absence of spatially varying power profiles. Phase~II will extend the CHF benchmark to include more complex geometries and boundary conditions, to be used along with transfer learning techniques, advanced UQ, and design optimization, all of which require a richer, more diverse CHF dataset.

To support these Phase II tasks, KSU, in collaboration with Westinghouse Electric Sweden, compiled and standardized published CHF measurements in tubes with uniform and non-uniform axial power distributions. This paper describes the resulting datasets; details on the original data collection can be found in the report \cite{bourisaw2025}.

\section{Data Collection and Organization}
\subsection{Data Source and Characteristics}
The primary source of the experimental data collected comes from the Korea Atomic Energy Research Institute (KAERI) technical report~\cite{park2000}. The report investigates how axial heat flux distribution affects the occurrence of CHF in boiling systems. The study includes an extensive set of experimental CHF measurements under both uniform and non-uniform heating conditions, providing valuable data for understanding and modeling CHF behavior in nuclear thermal-hydraulic systems. 

Overall, $1539$ CHF test cases were compiled and standardized from~\cite{park2000}, comprising $651$ uniform heating measurements and $888$ non-uniform heating cases, summarized in Table~\ref{tab:1}. The experimental setups used water coolant in vertical round tubes with internal diameter of $5.44$ -- $28.3$~mm and heated lengths of $0.061$ -- $7.0$~m.

\begin{table}[ht]
\centering
\captionsetup{justification=centering}
\caption{Summary of the standardized datasets.}
\setlength{\tabcolsep}{4pt}
\scriptsize
\begin{tabular}{|l|c|c|c|}\hline
\makecell[tc]{Heating} & \makecell[tc]{Pressure,\\MPa} & \makecell[tc]{Mass flux,\\kg/(m$^2\cdot$s)} & \makecell[tc]{Outlet equilibrium\\quality} \\\hline
Uniform    & 0.43 -- 18 & 335.0 -- 9561.9 & $-0.582$ -- 0.901 \\\hline
Non-uniform & 0.43 -- 18 & 328.2 -- 8916.0 & $-0.714$ -- 1.091 \\\hline
\end{tabular}
\label{tab:1}
\end{table}

\begin{figure}[ht]
    \centering
    \includegraphics[width=0.375\textwidth]{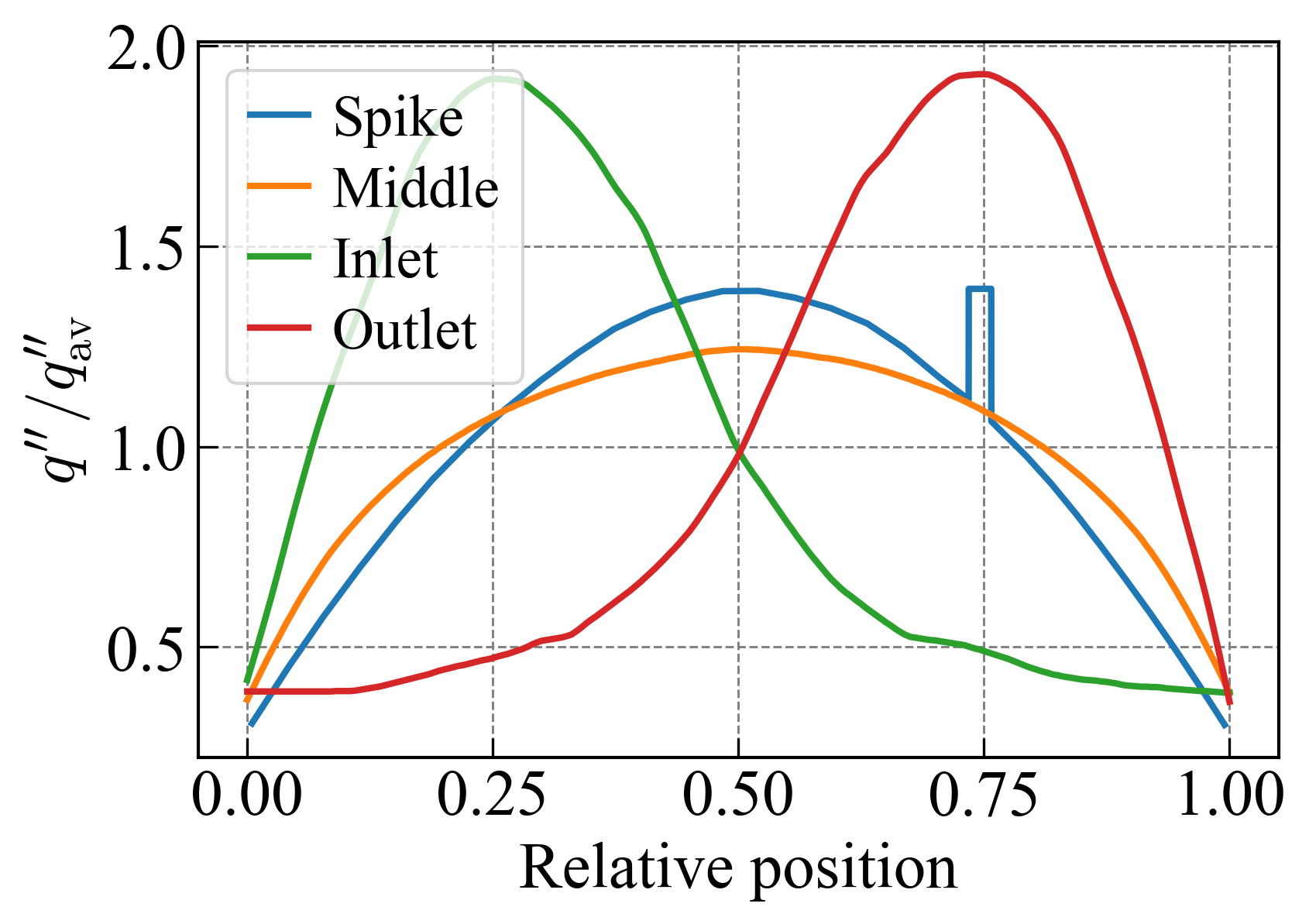}
    \captionsetup{justification=centerlast}
    \caption{Examples of the collected heat flux profiles. $q^{\prime\prime}$ denotes the heat flux, $q^{\prime\prime}_\text{av}$ denotes the average heat flux.}
    \label{fig:1}
\end{figure}

The non-uniform subset encompasses a variety of axial heat flux profiles. Representative examples of digitized profiles, adapted from one of the original sources~\cite{swenson1963}, are shown in Fig.\ref{fig:1}. These profiles are classified as ``spike'', ``middle-peaked'', ``inlet-peaked'', ``outlet-peaked'', or ``double-peaked''.

\subsection{Data Collection Workflow}
The uniform heating data were compiled from the KAERI report and checked against available original publications. For non-uniform heating, profiles were compiled from published figures and tabulated heating-zone data, using WebPlotDigitizer~\cite{rohatgi2019} where necessary, with the following steps:

\begin{compactenum}
    \item \emph{Outliers removal}: Spurious clicks in the published curves were manually discarded.
    \item \emph{Profile discretization}: Smooth heating profiles were represented by $40$ axial intervals, while stepwise profiles retained their original heating-zone divisions.
    \item \emph{Power normalization}: The heating profiles were normalized to a length-weighted mean of unity, and the average heat flux was calculated from total power and heated area. Source-based or compiled reference heat-flux values were preserved separately.
\end{compactenum}

For the uniform cases, inlet conditions and outlet equilibrium qualities were compiled from the KAERI report~\cite{park2000}, with documented corrections based on available original sources and thermodynamic calculations using IAPWS properties~\cite{iapws_python}. For uniform heating, the axial power distribution is represented by two equal intervals, while thermodynamic information is limited to the inlet conditions and outlet equilibrium quality. In the non-uniform dataset, the outlet equilibrium qualities were retained from the compiled or verified source data. Inlet values were generally retained, with extrapolation required for four Swenson spike cases, while intermediate values were reconstructed using the cumulative axial heating distribution.

\subsection{Dataset Structure}
The quantities were saved in separate (uniform and non-uniform) \texttt{XML} files which are easily readable by computer scripts, see Table~\ref{tab:2} that describes the saved quantities. The standardized \texttt{XML} files and accompanying documentation are available on Zenodo~\cite{bourisaw2025_data}. An additional uniform dataset containing 464 cases is provided after screening out 187 cases with confirmed or potential overlap with the NRC database. These datasets support CHF model development and benchmarking, including evaluation of axial power distribution effects.

\begin{table*}[ht]
\centering
\captionsetup{justification=centering}
\caption{Quantities saved for each test case.}
\setlength{\tabcolsep}{4pt}
\small
\begin{tabular}{|p{0.33\textwidth}|p{0.27\textwidth}|p{0.30\textwidth}|}\hline
\multicolumn{1}{|c|}{\textbf{Parameter}} & \textbf{Keyword} & \textbf{Units / Note} \\\hline
Test \# & <TestID> & Sequential (e.g., from 1 to 651 for uniform) \\\hline
Internal tube diameter & <Diameter> & [m] \\\hline
Wetted perimeter & <Perimeter> & [m] \\\hline
Flow area & <Area> & [m$^2$] \\\hline
Total heated length & <Length> & [m] \\\hline
Pressure & <Pressure> & [Pa] \\\hline
Total power & <Power> & [W] \\\hline
Mass flux & <MassFlux> & [kg/(m$^2\cdot$s)] \\\hline
Mass flow rate & <MassFlow> & [kg/s] \\\hline
Inlet temperature & <InletTemperature> & [$^\circ$C] \\\hline
Inlet enthalpy & <InletEnthalpy> & [J/kg] \\\hline
Average heat flux, $q^{\prime\prime}_\text{av}$ & <HeatFlux> & [W/m$^2$], total power / heated area \\\hline
Reported heat flux & <ReportedHeatFlux> & [W/m$^2$], source-dependent definition \\\hline
Reported heat flux definition & <ReportedHeatFluxDefinition> & [--], source convention \\\hline
Average critical heat flux & <AverageCriticalHeatFlux> & [W/m$^2$], identical to <HeatFlux> \\\hline
Local critical heat flux & <LocalCriticalHeatFlux> & [W/m$^2$], derived or \texttt{NaN} \\\hline
Heat flux definition & <HeatFluxDefinition> & [--], standardized convention \\\hline
Coolant & <Fluid> & Water only \\\hline
Original data source & <Source> & Original reference within \cite{park2000} \\\hline
Data provenance & <Provenance> & Source verification, corrections, and limitations \\\hline
Axial heating-profile multiplier & <WallPower> & [--], $N$ values \\\hline
Normalized local power $q^{\prime\prime}(z)/q^{\prime\prime}_\text{av}$ & <NormalizedWallPower> & [--], $N$ values \\\hline
Axial heating-interval length, $\Delta z$ & <WallMesh> & [m], $N$ values ($N=2$ for uniform heating) \\\hline
Outlet equilibrium quality & <EquilibriumQuality>, <XOUT> & [--], identical values \\\hline
Axial equilibrium quality, $x(z)$ & <Quality> & [--], $N+1$ values (non-uniform only) \\\hline
Axial coordinate, $z$ & <QualityPosition> & [m], $N+1$ values (non-uniform only) \\\hline
\multicolumn{3}{|c|}{\textbf{Additional Non-Uniform Entries}}\\\hline
Reported or assigned CHF coordinate & <CHFLocation> & [m], \texttt{NaN} if unavailable \\\hline
Power shape & <Shape> & Original KAERI designation \\\hline
Power shape category & <ShapeCategory> & ``inlet'', ``middle'', ``outlet'', ``spike'', or ``double-peaked'' \\\hline
Heating-profile continuity & <Continuous> & ``yes'' for smooth profiles; ``no'' for physical steps \\\hline
\end{tabular}
\label{tab:2}
\end{table*}

Not all compiled quantities could be independently verified. In the non-uniform dataset, CHF locations are unavailable for 36 cases, while local CHF values are indeterminate for 53 cases, including 17 with CHF locations at physical heating-step boundaries. These values are stored as \texttt{NaN}, and source-specific corrections and limitations are documented in the \texttt{Provenance} field. Identical or closely matched records are retained and should be grouped during data partitioning to prevent information leakage in ML applications.

\section{Preliminary Modeling Efforts}
The results presented in this section were obtained using the initial compiled datasets, before the corrections and standardization described above. These calculations have not been repeated using the standardized datasets.

\subsection{Traditional Correlations}
We first tested traditional CHF correlations, such as those by Bowring~\cite{bowring1972}:
\begin{equation*}
    q^{\prime\prime}_\text{cr} = \dfrac{A-B\,h_\text{fg}\,x}{C},
\end{equation*}
and Biasi~\cite{biasi1967}:
\begin{equation*}
    q^{\prime\prime}_\text{cr} = 15.048 \times 10^{7} (100\,D)^{-n}\,G^{-0.6}H\,(1-x),
\end{equation*}
where $A$, $B$, $C$ and $H$ are empirical parameters that depend on the thermodynamic state, flow conditions and geometric factors, $h_\text{fg}$ is the latent heat of vaporization, $x$ is the steam quality, $D$ is the hydraulic diameter, and $G$ is the mass flux.

\begin{figure}[ht]
\centering
\begin{subfigure}[t]{0.3\textwidth}
\centering
\includegraphics[width=\textwidth]{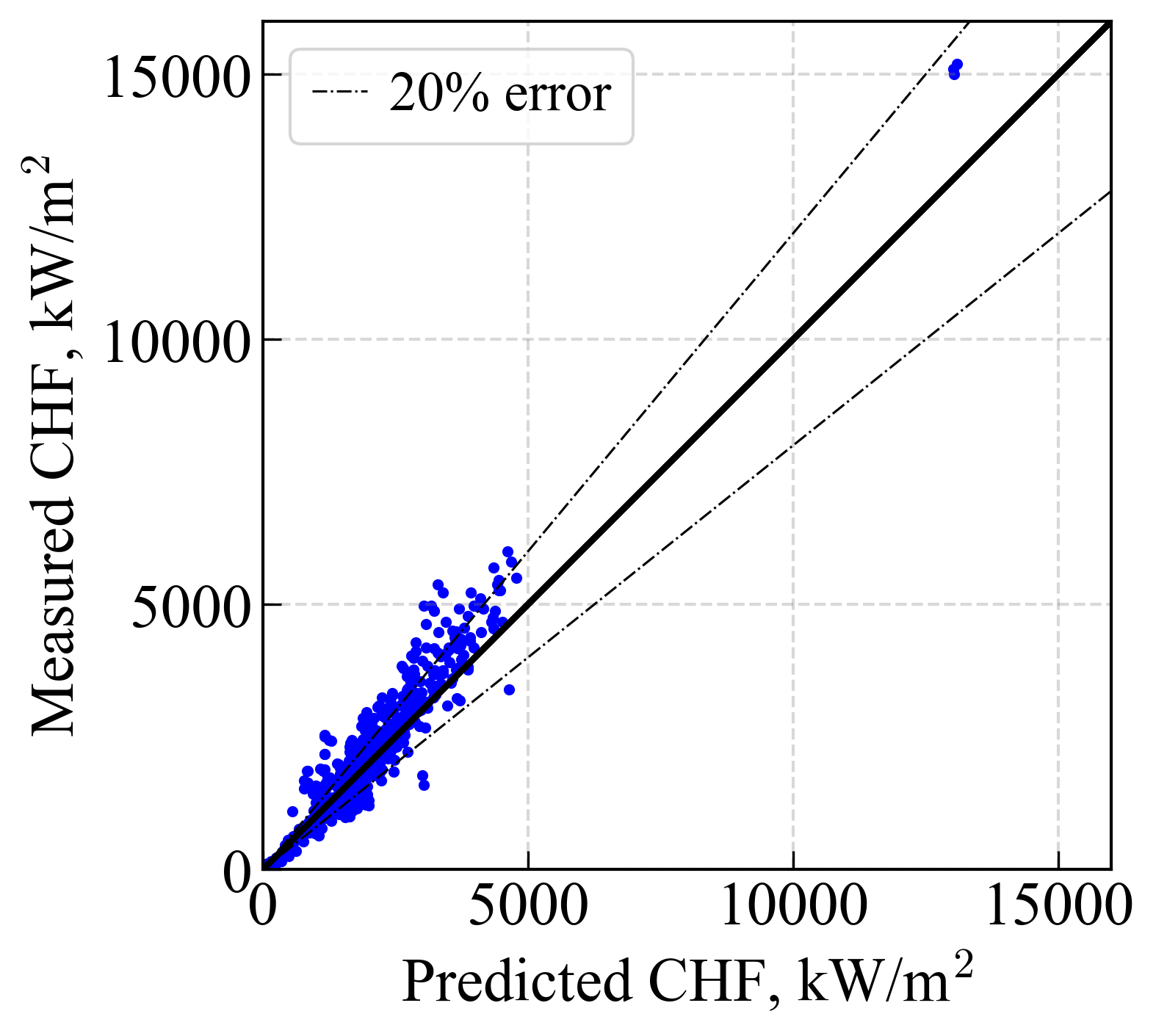}
\caption{}
\label{fig:2a}
\end{subfigure}
\\
\begin{subfigure}[t]{0.3\textwidth}
\centering
\includegraphics[width=\textwidth]{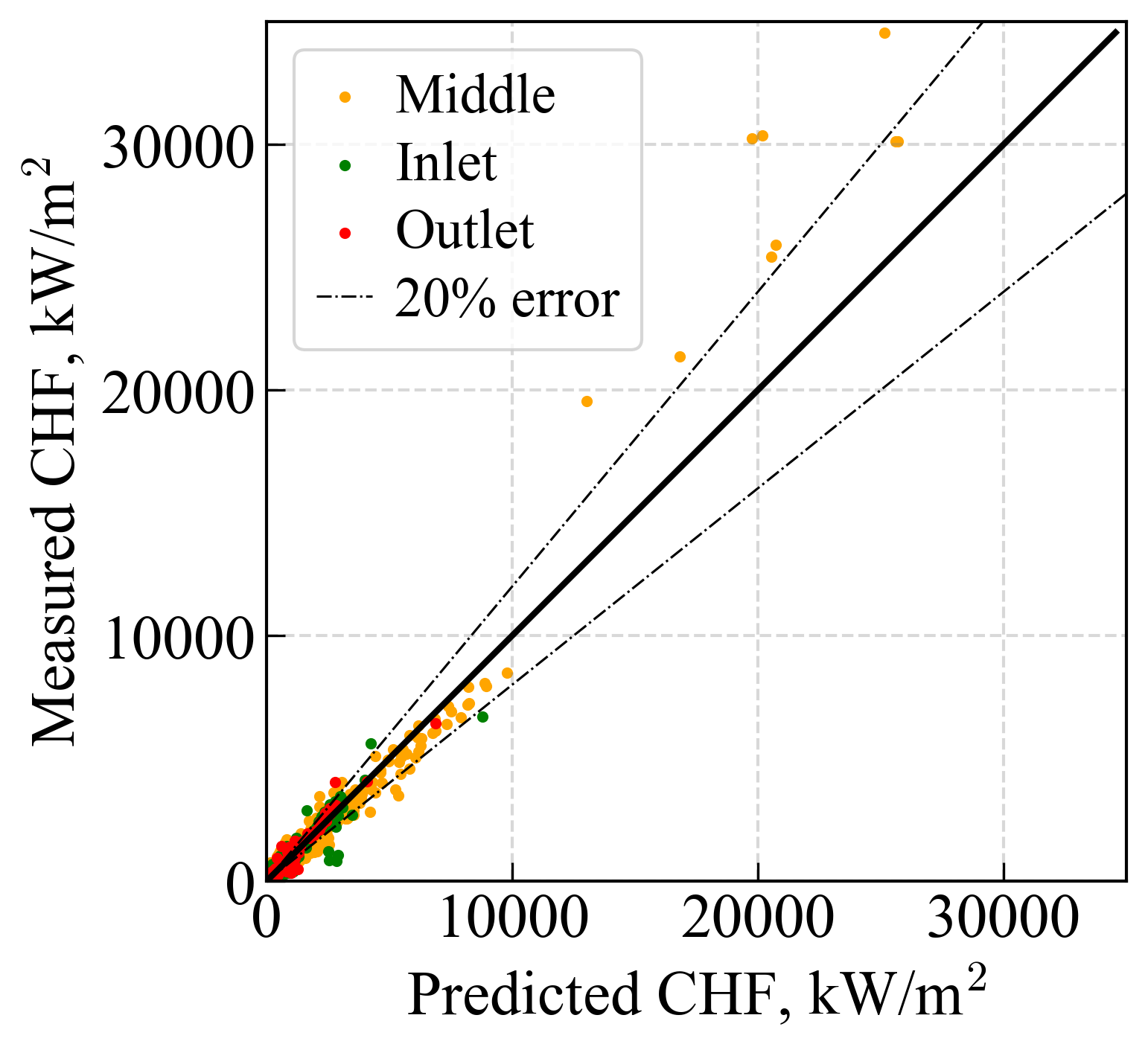}
\caption{}
\label{fig:2b}
\end{subfigure}
\captionsetup{justification=centering}
\caption{Predicted vs. measured CHF using LUT for (\subref{fig:2a})~uniform and (\subref{fig:2b})~non-uniform data.}
\label{fig:2}
\end{figure}

These correlations were used to predict the collected dataset and were evaluated separately for uniform and non-uniform heating conditions. For uniform cases, Bowring's correlation yields a root mean square error (RMSE) of 27\%, which increases significantly to 74\% for non-uniform cases. Biasi's correlation performs worse, with an RMSE of 62\% for uniform cases and 74\% for non-uniform ones. Both models occasionally produce unphysical (negative) CHF values, particularly near the limits of their applicability, and exhibit a systematic bias in predicting the CHF location under non-uniform heating. In contrast, the LUT~\cite{groeneveld2007}, including a non-uniform axial heat flux correction factor \cite{todreas2021}, demonstrates better overall performance, achieving a lower RMSE of approximately 20\% for uniform cases and maintaining greater robustness under non-uniform conditions, with an RMSE of 36\%, see Fig.~\ref{fig:2}.

\subsection{Neural Network Modeling}
\begin{figure}[ht]
\centering
\begin{subfigure}[t]{0.3\textwidth}
\centering
\includegraphics[width=\textwidth]{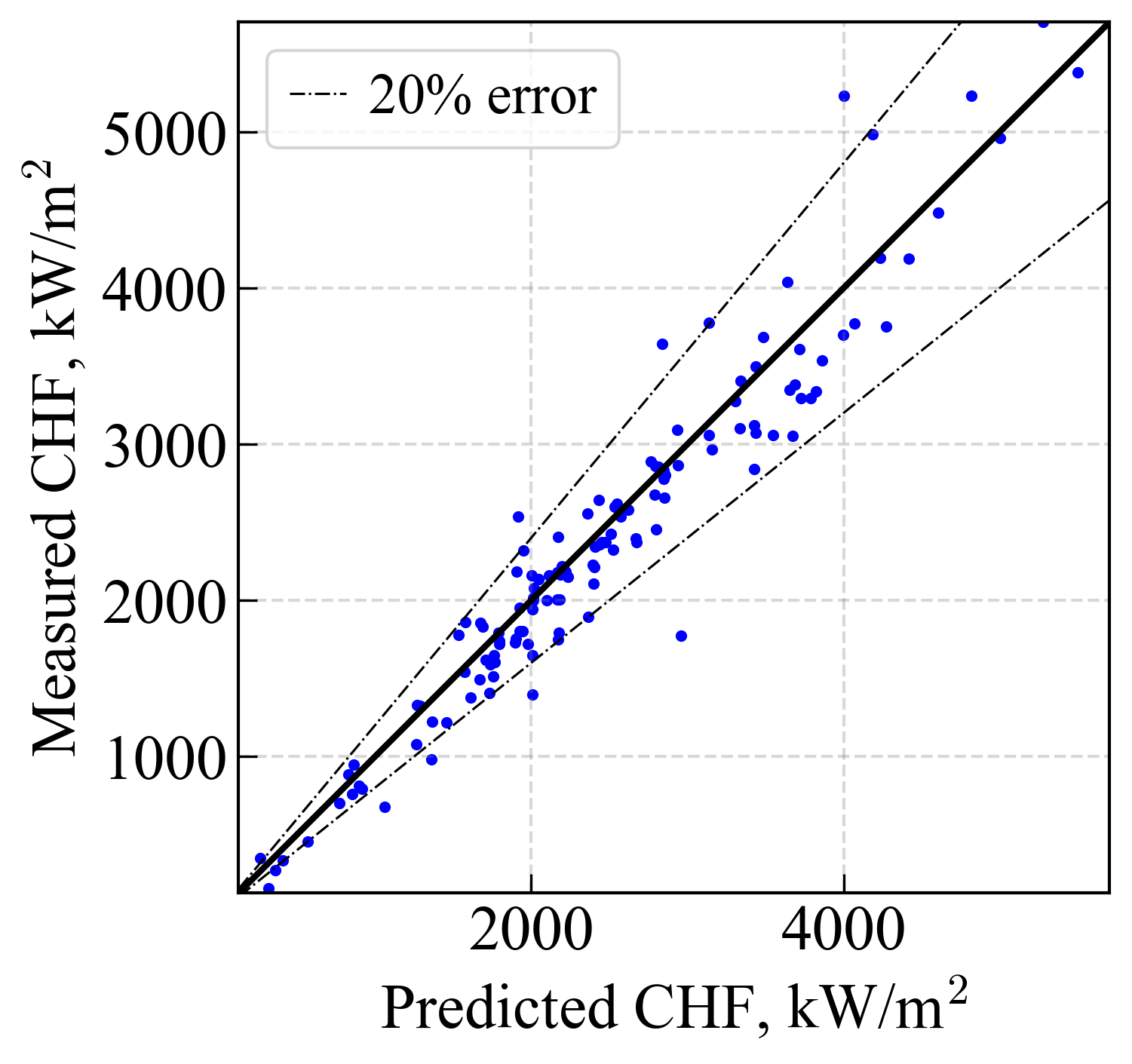}
\caption{}
\label{fig:3a}
\end{subfigure}
\\
\begin{subfigure}[t]{0.3\textwidth}
\centering
\includegraphics[width=\textwidth]{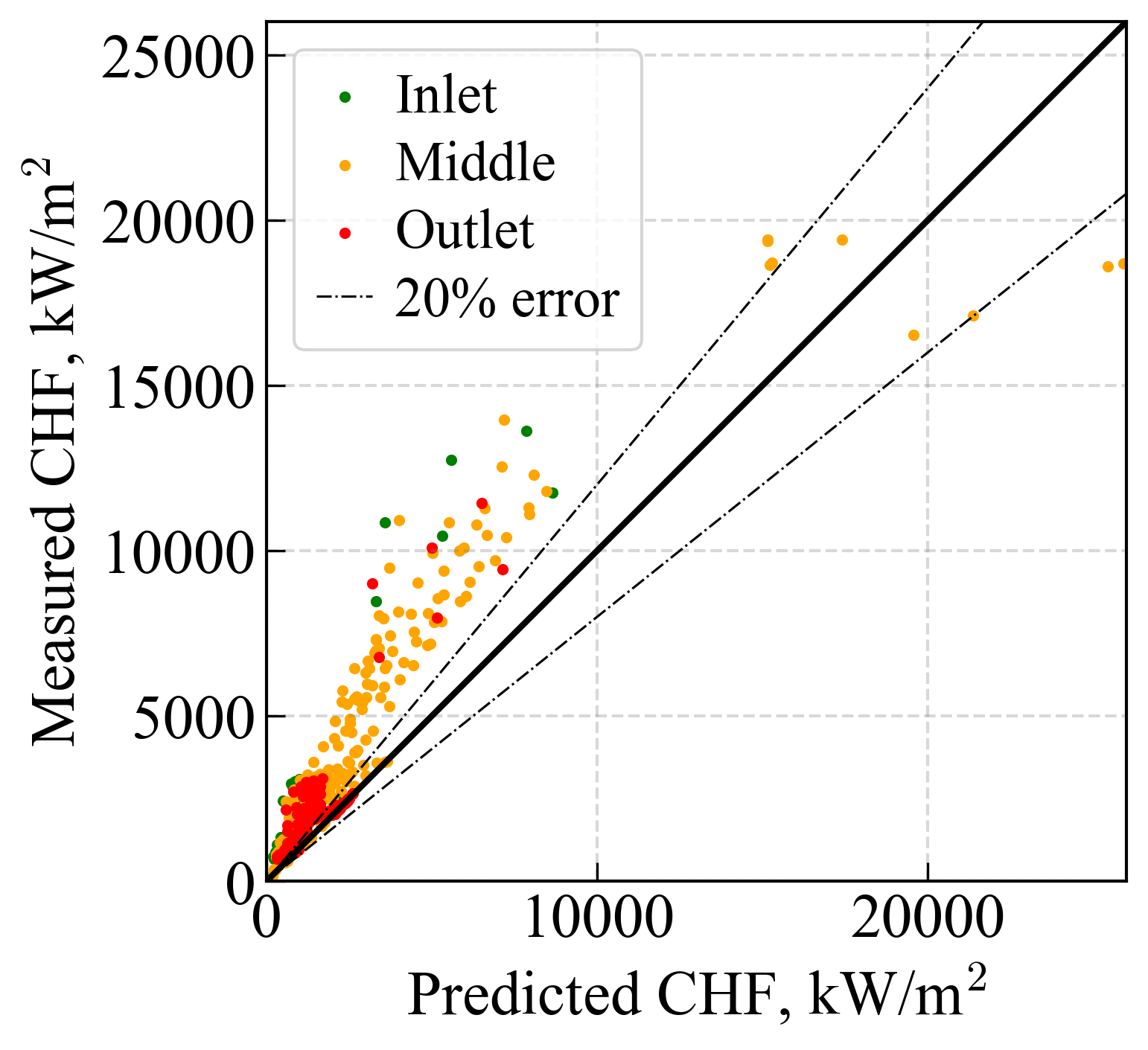}
\caption{}
\label{fig:3b}
\end{subfigure}
\captionsetup{justification=centering}
\caption{Predicted vs. measured CHF using NN for (\subref{fig:3a})~uniform and (\subref{fig:3b})~non-uniform data.}
\label{fig:3}
\end{figure}

A feedforward neural network (NN) was implemented following architecture by \cite{grosfilley2024}, see Table~\ref{tab:3}. The inputs comprised tube diameter, heated length, pressure, mass flux and equilibrium quality. The NN was trained to predict the CHF value on the uniform subset and achieved relatively high accuracy in predicting the held out uniform data. When this NN was used to predict the non-uniform CHF data, the accuracy significantly degraded, see Fig.~\ref{fig:3}. These results indicate the importance of accounting for axial heat flux distributions in CHF prediction, consistent with the objectives of Phase II of the CHF benchmark.

\begin{table}[ht]
\centering
\captionsetup{justification=centering}
\caption{Hyperparameters and NN architecture.}
\footnotesize
\begin{tabular}{|c|c|} \hline
\multicolumn{1}{|c}{\textbf{Parameter}} & \multicolumn{1}{|c|}{\textbf{Value / Method}} \\\hline
\makecell[tc]{Number of layers\\and neurons} & \makecell[tc]{5 -- 61 -- 51 -- 28 -- 39 --\\26 -- 21 -- 20 -- 14 -- 1} \\ \hline
Activation                   & ReLU                           \\ \hline
Optimizer                    & Adam                           \\ \hline
Loss function                & Mean squared error             \\ \hline
Regularization               & Dropout 1\% after the first layer \\ \hline
Learning rate                & 0.01 with decay on plateau     \\ \hline
\end{tabular}
\label{tab:3}      
\end{table}

\section{Conclusions}
In this study, we compiled and standardized a CHF dataset of $1539$ cases, including $651$ uniform and $888$ non-uniform heating experiments, from the KAERI technical report. The datasets were organized into separate \texttt{XML} files containing experimental conditions, thermodynamic quantities, axial heating profiles, and source-specific provenance information. Selected experimental values were verified and corrected, and heating profiles and heat-flux definitions were standardized. An additional uniform dataset containing $464$ cases was prepared to reduce potential overlap with the NRC database. Preliminary modeling using the initial compiled datasets showed that the LUT outperformed classical correlations, while an NN trained on uniform heating data exhibited reduced accuracy under non-uniform heating. The standardized datasets provide a documented basis for further CHF modeling and benchmarking activities within Phase II.

\section{Acknowledgments}
The authors would like to thank Dr. Mooneon Lee and Dr. Cheol Park of the KAERI for providing access to the report KAERI/TR-1665/2000.

\bibliographystyle{ans}
\bibliography{bibliography}

@book{todreas2021,
    author    = {Todreas, Neil E. and Kazimi, Mujid S.},
    title     = {Nuclear Systems Volume I: Thermal Hydraulic Fundamentals},
    publisher = {CRC Press},
    address   = {Boca Raton, FL},
    year      = {2021},
    edition   = {3rd},
}

@article{lee1988,
    title = {A Mechanistic Critical Heat Flux Model for Subcooled Flow Boiling Based on Local Bulk Flow Conditions},
    journal = {International Journal of Multiphase Flow},
    volume = {14},
    number = {6},
    pages = {711-728},
    year = {1988},
    issn = {0301-9322},
    doi = {https://doi.org/10.1016/0301-9322(88)90070-5},
    url = {https://www.sciencedirect.com/science/article/pii/0301932288900705},
    author = {C.H. Lee and I. Mudawar}
}

@article{guo2025,
  author    = {Guo, Yun and Zhang, Yijun},
  title     = {Review of Experimental and Numerical Studies on Critical Heat Flux in China},
  journal   = {Energies},
  year      = {2025},
  volume    = {18},
  number    = {4},
  pages     = {781},
  doi       = {10.3390/en18040781},
  url       = {https://doi.org/10.3390/en18040781},
}

@article{groeneveld1986,
    author = {D.C. Groeneveld and S.C. Cheng and T. Doan},
    title = {1986 AECL-UO Critical Heat Flux Lookup Table},
    journal = {Heat Transfer Engineering},
    volume = {7},
    number = {1-2},
    pages = {46--62},
    year = {1986},
    publisher = {Taylor \& Francis},
    doi = {10.1080/01457638608939644},
    URL = {https://doi.org/10.1080/01457638608939644},
    eprint = {https://doi.org/10.1080/01457638608939644}
}

@article{groeneveld2007,
    title = {The 2006 CHF Look-Up Table},
    journal = {Nuclear Engineering and Design},
    volume = {237},
    number = {15},
    pages = {1909-1922},
    year = {2007},
    note = {{NURETH-11}},
    issn = {0029-5493},
    doi = {https://doi.org/10.1016/j.nucengdes.2007.02.014},
    url = {https://www.sciencedirect.com/science/article/pii/S0029549307002002},
    author = {D.C. Groeneveld and J.Q. Shan and A.Z. Vasić and L.K.H. Leung and A. Durmayaz and J. Yang and S.C. Cheng and A. Tanase}
}

@article{grosfilley2024,
    author = {Grosfilley, Emil Helmryd and Robertson, Gustav and Soibam, Jerol and Le Corre, Jean-Marie},
    title = {Investigation of Machine Learning Regression Techniques to Predict Critical Heat Flux Over a Large Parameter Space},
    journal = {Nuclear Technology},
    pages = {1--15},
    year = {2024},
    publisher = {Taylor \& Francis},
    doi = {10.1080/00295450.2024.2380580},
    URL = {https://doi.org/10.1080/00295450.2024.2380580},
    eprint = { https://doi.org/10.1080/00295450.2024.2380580}
}

@techreport{lecorre2024,
      author       = {Le Corre, Jean-Marie and Delipei, Gregory and Wu, Xu and Zhao, Xingang},
      title        = {Benchmark on Artificial Intelligence and Machine Learning for Scientific Computing in Nuclear Engineering: Phase 1 Critical Heat Flux Exercise Specifications},
      institution  = {OECD Nuclear Energy Agency},
      number       = {NEA/WKP(2023)1},
      month        = {January},
      year         = {2024},
      url          = {https://www.oecd-nea.org/upload/docs/application/pdf/2024-01/nea_wkp_2023_1_2024-01-18_16-23-40_237.pdf},
}

@techreport{park2000,
  author       = {Park, Cheol},
  title        = {Effect of Axial Heat Flux Distribution on Critical Heat Flux},
  institution  = {Korea Atomic Energy Research Institute},
  number       = {KAERI/TR-1665/2000},
  address      = {Daejeon, Republic of Korea},
  month        = {October},
  year         = {2000},
  url          = {https://inis.iaea.org/records/xygaz-3k954}
}

@techreport{bowring1972,
  author       = {Bowring, R.~W.},
  title        = {Simple but Accurate Round Tube, Uniform Heat Flux, Dryout Correlation over the Pressure Range 0.7 to 17\,MN/m\textsuperscript{2} (100 to 2500\,psia)},
  institution  = {Atomic Energy Establishment, Winfrith, England},
  type         = {Technical Report},
  number       = {AEEWR789},
  year         = {1972},
  pages        = {18},
  note         = {NTIS Issue Number 197223},
  url          = {https://ntrl.ntis.gov/NTRL/dashboard/searchResults/titleDetail/AEEWR789.xhtml},
}

@article{biasi1967,
  author  = {Biasi, L. and Clerici, G.~C. and Garribba, S. and Sala, R. and Tozzi, A.},
  title   = {Studies on Burnout. Part 3. A New Correlation for Round Ducts and Uniform Heating and Its Comparison with World Data},
  journal = {Energ. Nucl. (Milan)},
  volume  = {14},
  pages   = {530--536},
  month   = {September},
  year    = {1967},
  url     = {https://www.osti.gov/biblio/4223933},
  note    = {OSTI ID: 4223933},
}

@misc{rohatgi2019,
  author       = {Rohatgi, Ankit},
  title        = {WebPlotDigitizer},
  howpublished = {https://automeris.io/WebPlotDigitizer},
  note         = {Version 4.2; accessed June 16, 2025},
  year         = {2019}
}

@inproceedings{swenson1963,
  author    = {Swenson, H. S. and Carver, J. R. and Kakarala, C. R.},
  title     = {The Influence of Axial Heat Flux Distribution on the Departure from Nucleate Boiling in a Water Cooled Tube},
  booktitle = {ASME Winter Annual Meeting},
  year      = {1962},
  number    = {62-WA-297},
  organization = {ASME}
}

@misc{iapws_python,
  author       = {José Joaquín Gómez-Romero},
  title        = {IAPWS: Python Implementation of IAPWS-IF97 and Other Standards},
  year         = {2024},
  howpublished = {https://pypi.org/project/iapws/},
  note         = {Accessed: 2025-06-16}
}

@techreport{bourisaw2025,
  title        = {Critical Heat Flux Prediction and Mitigation in Light Water Reactors},
  author       = {Bourisaw, Reece and McCants, Reid and Le Corre, Jean-Marie and Iskhakov, Arsen S.},
  institution  = {Kansas State University, Department of Mechanical and Nuclear Engineering},
  type         = {Technical Report},
  number       = {},
  month        = jun,
  year         = {2025},
  doi          = {10.13140/RG.2.2.30080.96000},
  url          = {https://www.researchgate.net/publication/392786992_Critical_Heat_Flux_Prediction_and_Mitigation_in_Light_Water_Reactors},
  note         = {Accessed: 2025-06-18}
}

@misc{bourisaw2025_data,
  author       = {Bourisaw, Reece and McCants, Reid and Le Corre, Jean-Marie and Iskhakova, Anna and Iskhakov, Arsen S.},
  title        = {Critical Heat Flux Datasets under Uniform and Non-Uniform Axial Heating Conditions [KAERI/TR-1665/2000]},
  year         = {2026},
  publisher    = {Zenodo},
  note         = {Version 3, October 22, 2026},
  doi          = {10.5281/zenodo.23566164},
  url          = {https://doi.org/10.5281/zenodo.23566164}
}

\end{document}